\documentclass[sigconf]{acmart}

\pdfoutput=1 

\renewcommand\footnotetextcopyrightpermission[1]{} 
\usepackage{mathtools}
\usepackage{multirow}
\usepackage[ruled,vlined]{algorithm2e}
\usepackage{graphicx}
\usepackage{caption}
\usepackage{subcaption}
\usepackage{amsmath}
\usepackage{setspace}
\usepackage{hyperref}
\usepackage{color}
\usepackage{multirow}
\usepackage{booktabs}
\newcommand{\norm}[1]{\left\lVert#1\right\rVert}

\DeclareMathOperator*{\argmax}{\operatornamewithlimits{argmax}}

\usepackage{booktabs} 

\begin{document}

\title{Learning a Neural-network-based Representation for Open Set Recognition}

\author{Mehadi Hassen}
\affiliation{%
  \institution{School of Computing, Florida Institute of Technology}
  \streetaddress{150 West University Blvd}
  \city{Melbourne}
  \state{FL 32901}
}
\email{mhassen2005@my.fit.edu}

\author{Philip K. Chan}
\orcid{0000-0002-3878-4205}
\affiliation{%
  \institution{School of Computing, Florida Institute of Technology}
  \streetaddress{150 West University Blvd}
  \city{Melbourne}
  \state{FL 32901}
}
\email{pkc@cs.fit.edu}

\begin{abstract}
Open set recognition problems exist in many domains. For example in security, new malware classes emerge regularly; therefore malware classification systems need to identify instances from unknown classes in addition to discriminating between known classes. In this paper we present a neural network based representation for addressing the open set recognition problem. In this representation instances from the same class are close to each other while instances from different classes are further apart, resulting in statistically significant improvement when compared to other approaches on three datasets from two different domains.    

\end{abstract}

%
%



\maketitle

\section{Introduction}
\label{sec:intro}

To build robust AI systems, Dietterich \cite{dietterich2017steps} reasons that one of the main challenges is handling the ``unknown unknowns."  One idea is to detect model failures; that is, the system understands that its model about the world/domain has limitations and may fail. For example, assume you trained a binary classifier model to discriminate between pictures of cats and dogs. You deploy this model and observe that it does a very good job at recognizing images of cats and dogs. What would this model do if it is faced with a picture of a fox or a caracal (mid sized African wild cat)? The model being a binary classifier will predict these pictures to be either a dog or a cat, which is not desirable and can be considered as a failure of the model. In machine learning, one direction of research for detecting model failure is ``open category learning", where not all categories are known during training and the system needs to appropriately handle instances from novel/unknown categories that may appear during testing. Besides ``open category learning", terms such as  ``open world recognition" \cite{bendale2015towards} and ``open set recognition" \cite{scheirer2013toward,bendale2016towards} have been used in past literatures. In this paper we will use the term ``open set recognition". 

Where does open set recognition appear in real world problems? There are various real world applications that operate in an open set scenario. For example, Ortiz and Becker \cite{ortiz2014face} point to the problem of face recognition. One such use case is automatic labeling of friends in social media posts, ``where the system must determine if the query face exists in the known gallery, and, if so, the most probable identity." Another domain is in malware classification, where training data usually is incomplete because of novel malware families/classes that emerge regularly. As a result, malware classification systems operate in an open set scenario.

In this paper we propose a neural network based representation and a mechanism that utilizes this representation for performing open set recognition. Since our main motivation when developing this approach was the malware classification domain, we evaluate our work on two malware datasets. To show the applicability of our approach to domains outside malware, we also evaluate our approach on images. 

Our contributions include: (1) we propose an approach for learning a representation that facilitates open set recognition, (2) we propose a loss function that enables us to use the same distance function both when training and when computing an outlier score, (3) our proposed approaches achieve statistically significant improvement compare to previous research work on three datasets.

The remainder of this paper is organized in the following manner. In Section \ref{sec:related_works}, we give an overview of related works and present our approach in Section \ref{sec:approach}. We present our evaluation methodology, results, and provide further discussions in Section \ref{sec:eval}.

\section{Related Work}
\label{sec:related_works}

We can broadly categorize existing open set recognition systems into two types. The first type provides mechanisms to discriminate known class instances from unknown class instances. These systems, however, cannot discriminate between the known classes, where there is more than one. Research works such as  \cite{scheirer2013toward,bodesheim2013kernel,bodesheim2015local} fall in this category. Scheirer et al. \cite{scheirer2013toward} formalized the concept of open set recognition and proposed a 1-vs-set binary SVM based approach. Bodesheim et al. \cite{bodesheim2013kernel} propose KNFST for performing open set recognition for multiple known classes at the same time. The idea of KNFST is further extended in \cite{bodesheim2015local} by considering the locality of a sample when calculating its outlier score. 

The second type of open set recognition system, provides the ability to discriminate between known classes in addition to identifying unknown class instances. Research works such as  \cite{Jain2014,bendale2015towards,bendale2016towards,da2014learning,ge2017generative} fall in this category. PI-SVM \cite{Jain2014}, for instance, uses a collection of binary SVM classifiers, one for each class, and fits a Weibull distribution over the score each classifier. This approach allows PI-SVM to be able to both perform recognition of unknown class instances and classification between the known class instances. Bendale and Boult \cite{bendale2015towards} propose an approach to extend Nearest Class Mean (NCM) to perform openset recognition with the added benefit of being able to do incremental learning. 

Neural Net based methods for open set recognition have been proposed in \cite{bendale2016towards, cardoso2015bounded, ge2017generative}. Openmax \cite{bendale2016towards} (a state-of-art algorithm) modifies the normal Softmax layer of a neural network by redistributing the activation vector (i.e. the values of the final layer of a neural network that are given as input to the Softmax function) to account for unknown classes. A Neural Net is first trained with the normal Softmax layer to minimize cross entropy loss. The activation vector of each training instance is then computed; and using these activation vectors the per-class mean of the activation vector (MAV) is calculated. Then each training instance's distance from its class MAV is computed and a separate Weibull distribution for each class is fit on certain number of the largest such distances. Finally, the activation vector's values are redistributed based on the probabilities from the Weibull distribution and the redistributed value is summed to represent the unknown class activation value. The class probabilities (now including the unknown class) are then calculated using Softmax on the new redistributed activation vector. 

The challenge of using the distance from MAV is that the normal loss functions, such as cross entropy, do not directly incentivize projecting class instances around the MAV. In addition to that,  because the distance function used during testing is not used during training it might not necessarily be the right distance function for that space. We address this limitation in our proposed approach.

Ge et al. \cite{ge2017generative} combine Openmax and GANs\cite{goodfellow2014generative} for open set recognition. The network used for their approach is trained on the known class instances plus synthetic instances generated using the DCGAN\citep{radford2015unsupervised}. In the malware classification domain, K. Rieck et al.\cite{rieck2011automatic} proposed a malware clustering approach and an associated outlier score. Although the authors did not propose their work for open set recognition, their outlier score can be used for unsupervised open set recognition. Rudd et al. \cite{rudd2017survey} outline ways to extend existing closed set intrusion detection approaches for open set scenarios.

\section{Approach}
\label{sec:approach}

For open set recognition, given a set of instances belonging to known classes, we would like to learn a function that can accurately classify an unseen instance to one of the known classes or an unknown class. Let $D$ be a set of instances $X$ and their respective class labels $Y$ (ie, $D = (X,Y)$), and $K$ be the number of unique known class labels.  Given $D$ for training, the problem of open set recognition is to learn a function $f$ that can accurately classify an unseen instance (not in $X$) to one of the $K$ classes or an unknown class (or the ``none of the above" class).

The problem of open set recognition differs from the problem of closed set (``regular") classification because the learned function $f$ needs to handle unseen instances that might belong to classes that are not known during training.  That is, the learner is robust in handling instances of classes that are not known.  This difference is the main challenge for open set recognition.  Another challenge is how to learn a more effective instance representation that facilitates open set recognition than the original instance representation used in $X$.

\subsection{Overview}
\label{sec:overview}
Consider $\vec{x}$ is an instance and $y = f(\vec{x})$ is the class label predicted using $f(\vec{x})$. In case of a closed set, $y$ is one of the known class labels. In the case of open set, $y$ could be one of the known classes or an unknown class. The hidden layers in a neural network, $\vec{z} = g(\vec{x})$, can be considered as different representations of $\vec{x}$. Note, we can rewrite $y$ in terms of the hidden layer as $y =  f(\vec{z}) = f(g(\vec{x}))$.

The objective of our approach is to learn a representation that facilitates open set recognition. We would like this new representation to have two properties: (P1) instances of the same class are closer together and (P2) instances of different classes are further apart.  The two properties can lead to larger spaces among known classes for instances of unknown classes to occupy.  Consequently, instances of unknown classes could be more effectively detected. 

This representation is similar in spirit to a Fisher Discriminant. A Fisher discriminant aims to find a linear projection that maximizes between class (inter class) separation while minimizing within class (intra class) spread. Such a projection is obtained by maximizing the Fisher criteria. However, in the case of this work, we use a neural network with a \textit{non-linear} projection to learn this representation.


For closed set classification, the hidden layer representations are learned to minimize classification loss of the output $y$ using loss functions such as cross entropy.  However, the representations might not have the two desirable properties we presented earlier, useful in open set recognition. In this paper we present an approach in Section \ref{sec:learning_representation} for learning the hidden layer $\vec{z} = g(\vec{x})$ using a loss function, that directly tries to achieve the two properties. 

Once such representation is learned, we can use the distance of $\vec{z}$ from a class center as an outlier score (i.e. further away $\vec{z}$ is from the closest class center the more likely it is to be from unknown class/outlier.) This is discussed in more detail in Section \ref{sec:outlier_score}. By observing how close training instances are to the class center, we estimate a threshold value for the outlier score for identifying unknown class instances (Section \ref{sec:threshold}). We can further use the distance of $\vec{z}$ from all the known class centers to predict a probability distribution over all the known classes (Section \ref{sec:class_prob}). 

To make the final open set prediction, we use the threshold on the outlier score to first identify known class instances from unknown class instances.  Then, for the instances that are predicted as known, we use the predicted probability distribution over the known classes to identify the most likely known class label (Section \ref{sec:open_set_classification}). 

\subsection{Learning representations}
\label{sec:learning_representation}
Recall that to learn a new representation $\vec{z}$ for an original instance $\vec{x}$, we learn a function $g$ that projects $\vec{x}$ to $\vec{z}$ (ie, $\vec{z}=g(\vec{x})$). The learning process is guided by a loss function that satisfies properties P1 and P2 in Section \ref{sec:overview}.

\subsubsection{II-Loss Function}
\label{sec:ii_loss}

In a typical neural network classifier, the activation vector that comes from the final linear layer are given as input to a Softmax function. Then the network is trained to minimize a loss function such as cross entropy on the outputs of the Softmax layer. In our case the output vector $\vec{z_i}$ of the final linear layer (i.e activation vector that serves as input to a softmax in a typical neural net) are considered as the projection of the input vector $\vec{x_i}$, of instance $i$, to a different space in which we aim to maximize the distance between different classes (inter class separation) and minimize distance of an instance from its class mean (intra class spread). We measure intra class spread as the average distance of instances from their class means:
\begin{equation}
intra\_spread=  \frac{1}{N} \sum_{j=1}^{K} \sum_{i=1}^{|C_j|}\norm{\vec{\mu_j} - \vec{z_i}}^2_2
\label{eq:intra_class}
\end{equation}
\noindent where $|C_j|$ is the number of training instances in class $C_j$, $N$ is the number of training instances, and $\mu_j$ is the mean of class $C_j$ :
\begin{equation}
\vec{\mu_j}= \frac{1}{|C_j|}\sum_{i=1}^{|C_j|}\vec{z_i}
\label{eq:class_mean}
\end{equation}

We measure the inter class separation in terms of the distance between the closest two class means among all the $K$ known classes: 
\begin{equation}
inter\_sparation= \min_{\substack{\mathllap{1\le} m \mathrlap{\le K} \\ \\ \mathllap{m+1 \le} n \mathrlap{\le K}}} \quad  \norm{\vec{\mu_m} - \vec{\mu_n}}^2_2
\label{eq:inter_class}
\end{equation}

\noindent An alternative for measuring inter class separation would have been to take the average distances between all class means. Doing so, however, allows the largest distance between two classes to dominate inter class separation, hence does not result in a good separation. 

The network is then trained using mini-batch stochastic gradient descent with backpropagation as outlined in Algorithm \ref{algo:training_with_ii} to minimize the loss function in Equation \ref{eq:ii_loss}, which we will refer to ii-loss for the remainder of this paper. This loss function minimizes the intra class spread and maximizes inter class separation.
\begin{equation}
\text{\textit{ii-loss}} = intra\_spread - inter\_sparation
\label{eq:ii_loss}
\end{equation}

\footnotesize
\begin{algorithm}[t]
  \LinesNumbered
  \SetKwData{D}{$D$}
  \SetKwData{X}{$X$}
  \SetKwData{Y}{$Y$}
  \SetKwData{Xbatch}{$X_{batch}$}
  \SetKwData{Ybatch}{$Y_{batch}$}
  \SetKwData{Zbatch}{$Z_{batch}$}
  \SetKwData{ClassMeans}{$\{\vec{\mu}_1 \cdots \vec{\mu}_{K}\}$}
  \SetKwData{L}{ii-loss}
  \SetKwFunction{CMean}{$compute\_class\_means$}
  \SetKwFunction{Intra}{$compute\_intra\_spread$}
  \SetKwFunction{Intraspread}{$intra\_spread$}
  \SetKwFunction{Inter}{$compute\_inter\_separation$}
  \SetKwFunction{Interseparation}{$inter\_separation$}
  \SetKwInOut{Input}{Input}
  \SetKwInOut{Output}{Output} 
  \Input{\\
  (\X, \Y): Training data and labels \\
  }
  \BlankLine
  \For{number of training iterations}{
  	 Sample a mini-batch ($X_{batch}$, $Y_{batch}$) from (\X, \Y) \\
     \Zbatch $\leftarrow$ $g(X_{batch})$ \\
     \ClassMeans $\leftarrow$ $class\_means(Z_{batch}, Y_{batch})$ \\
     \Intraspread $\leftarrow$ $intra\_spread(Z_{batch}, \{\vec{\mu}_1 \cdots \vec{\mu}_{K}\})$ \\
     \Interseparation $\leftarrow$ $inter\_separation(\{\vec{\mu}_1 \cdots \vec{\mu}_{K}\})$ \\
     \L $\leftarrow$ \Intraspread - \Interseparation \\
     update parameters of $g$ using stochastic gradient descent to minimize \L\\
  }
  \ClassMeans $\leftarrow$ $class\_means(g(X), Y)$ \\
  return \ClassMeans and parameters of $g$ as the model.
  
\caption{Training to minimize ii-loss.}
\label{algo:training_with_ii}
\end{algorithm}
\normalsize

After the network finishes training, the class means are calculated for each class using all the training instances of that class and stored as part of the model.

\begin{figure*}[ht]
        \centering
        \begin{subfigure}[b]{0.3\textwidth}
                \centering
                \includegraphics[scale=0.5]{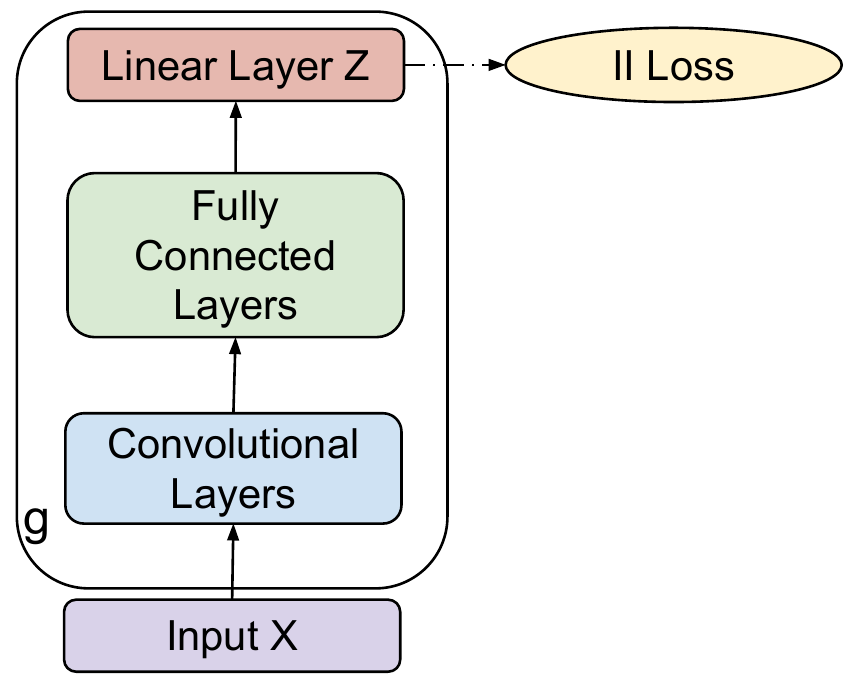}
                \caption{Convolutional network with ii-loss}
                \label{fig:con_ii_loss}
        \end{subfigure}
        ~ 
        \begin{subfigure}[b]{0.36\textwidth}
                \centering
                \includegraphics[scale=0.5]{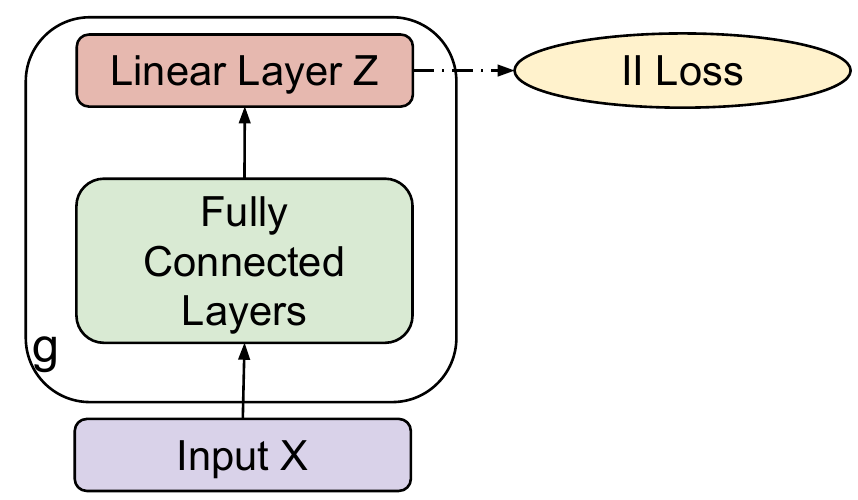}
                \caption{Fully connected network with ii-loss}
                \label{fig:flat_ii_loss}
        \end{subfigure}
        \begin{subfigure}[b]{0.3\textwidth}
                \centering
                \includegraphics[scale=0.5]{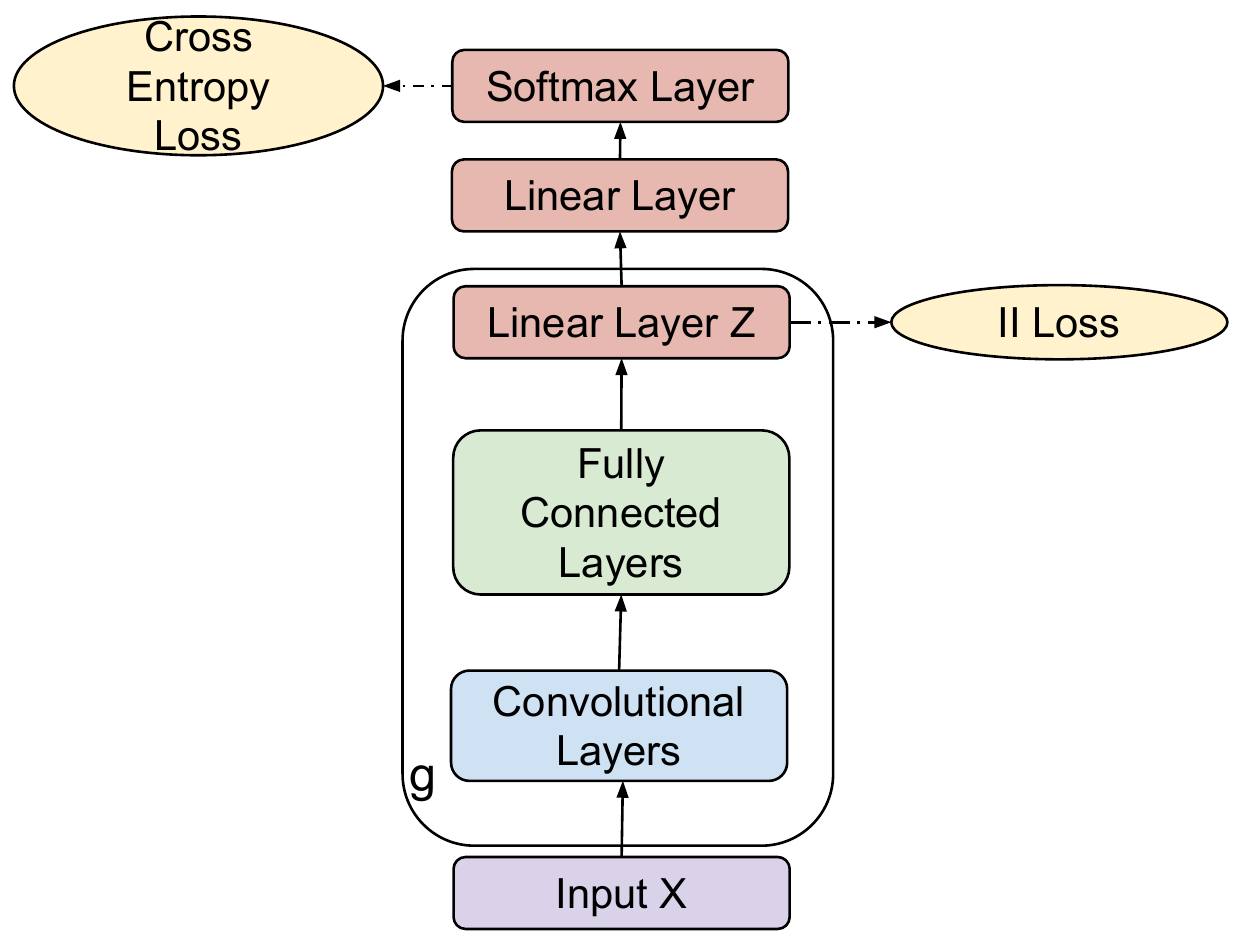}
                \caption{Combining ii-loss with Cross Entropy Loss}
                \label{fig:ceii_network}
        \end{subfigure}
        \caption{Network architecture with ii-loss.}
        \label{fig:network_ack}
\end{figure*}

The neural network $g$ used to learn the representation can be either a combination of convolution and fully connected layers, as shown in Figure \ref{fig:con_ii_loss}, or it can be all fully connected layers, Figure \ref{fig:flat_ii_loss}. Both types are used in our experimental evaluation. 

\subsubsection{Combining ii-loss with Cross Entropy Loss}
\label{sec:iice_loss}
While the two desirable properties P1 and P2 discussed in Section \ref{sec:overview} aim to have a representation that separates instances from different classes, lower classification error is not explicitly stated. Hence, a third desirable property (P3) is a low classification error in the training data. To achieve this, alternatively a network can be trained on both cross entropy loss and ii-loss (Eq \ref{eq:ii_loss}) simultaneously. The network architecture in Figure \ref{fig:ceii_network} can be used. In this configuration, an additional linear layer is added after the z-layer. The output of this linear layer is passed through a Softmax function to produce a distribution over the known classes. Although Figure \ref{fig:ceii_network} shows a network with a convolutional and fully connected layers, combining ii-loss with cross entropy can also work with a network of fully connected layers only.


The network is trained using mini-batch stochastic gradient descent with backpropagation. During each training iterations the network weights are  first updated to minimize on ii-loss and then updated to minimize cross entropy loss. Other researchers have trained neural networks using more than one loss function. For example, the encoder network of an Adversarial autoencoders \cite{makhzani2015adversarial} is updated both to minimize the reconstruction loss and the generators loss.

\subsection{Outlier Score for Open Set Recognition}
\label{sec:outlier_score}

During testing we use an outlier score to indicate the degree to which the network predicts an instance $\vec{x}$ to be an outlier. This outlier score is calculated as the distance of an instance to the closest class mean from among $K$ known classes.
\begin{equation}
outlier\_score(\vec{x}) = \min_{1 \le j \le K} \norm{\vec{\mu}_j - \vec{z}}^2_2
\label{eq:outlier_score}
\end{equation}
\noindent where $\vec{z} = g(\vec{x})$.

Because the network is trained to project the members of a class as close to the mean of the class as possible the further away the projection $\vec{z}$ of instance $\vec{x}$ is from the closest class mean, the more likely the instance is an outlier for that class. 

\subsubsection{Threshold Estimation}
\label{sec:threshold}

Once a proper score is identified to be used as an outlier score, the next step is determining above which threshold value of this score will indicate an outlier. In other words, how far does the projection of an instance need to be from the closest class mean for it to be deemed an outlier. For this work, we propose a simple threshold estimation. To pick an outlier threshold, we assume that a certain percent of the training set to be noise/outliers. We refer to this percentage as the contamination ratio. For example, if we set the contamination ratio to be 0.01 it will be like assuming 1\% of the training data to be noise/outliers. Then, we calculate the outlier score on the training set instances, sort the scores in ascending order and pick the 99 percentile outlier score value as the outlier threshold value.


  

The reader might notice that the threshold proposed in this section is a global threshold. This means that the same outlier threshold value is used for all classes. An alternative to this approach is to estimate the outlier threshold per-class. However, in our evaluation we observe that global threshold consistently gives more accurate results than per-class threshold.

\subsection{Prediction on Known Classes}
\label{sec:class_prob}

Once a test instance is predicted as known, the next step is to identify to which of the known class it belongs. An instance is classified as belonging to the class whose class mean its projection is nearest to. If we want the predicted class probability over the known classes, we can take the softmax of the negative distance of a projection $\vec{z}$, of the test instance $\vec{x}$ (i.e. $\vec{z} = g(\vec{x})$), from all the known class means. Hence the predicted class probability for class $j$ is given by:
\begin{equation}
P(y=j \mid \vec{x}) = \frac{e^{-\norm{\vec{\mu}_j - \vec{z}}^2_2}}{\sum_{m=1}^{K} e^{-\norm{\vec{\mu}_m - \vec{z}}^2_2}} 
\label{eq:pred_prob}
\end{equation}

\subsection{Performing Open Set Recognition}
\label{sec:open_set_classification}

Open set recognition is a classification over $K + 1$ class labels, where the first $K$ labels are from the known classes the classifier is trained on, and the $K + 1$st label represents the $unknown$ class that signifies that an instance does not belong to any of the known classes. This is performed using the outlier score in Equation \ref{eq:outlier_score} and the $threshold$ estimated in Section \ref{sec:threshold}. The $outlier\_score$ of a test instance is first calculated. If the score is greater than $threshold$, the test instance is labeled as $K + 1$, which in our case corresponds to the \textit{unknown class}, otherwise the appropriate class label is assigned to the instance from among the known classes:
\begin{equation}
  y=\left\{
  \begin{array}{@{}ll@{}}
  	K + 1, & \text{if}\ outlier\_score > threshold \\
  	\argmax\limits_{1 \le j \le K}  P(y=j \mid \vec{x}), & \text{otherwise}
  \end{array}\right.
\label{eq:open_pred}
\end{equation} 

When a network is trained on ii-loss alone, $P(y=k \mid \vec{x})$ in the above equation comes from Equation \ref{eq:pred_prob}; whereas in case of a network trained on both ii-loss and cross entropy loss, discussed in Section \ref{sec:iice_loss}, $P(y=k \mid \vec{x})$ is from the Softmax layer in Figure \ref{fig:ceii_network}.

\section{Evaluation}
\label{sec:eval}

\begin{table*}
\centering
\caption{Average AUC of 30 runs up to 100\% FPR and 10\% FPR (the positive label represented instances from unknown classes and the negative label represented instances from the known classes when calculating the AUC). The \underline{underlined} average AUC values are higher with statistical significance (p-value < 0.05 with a t-test) compared to the values that are not underlined on the same row. The average AUC values in \textbf{bold} are the largest average AUC values in each row.}
\label{table:auc}
\begin{tabular}{l|l|l|l|l}
\toprule
                               & FPR   & ce                   & ii                   					  & ii+ce  \\ 
\midrule
\multirow{2}{*}{MNIST}         & 100\% & 0.9282 ($\pm$0.0179) & \underline{\textbf{0.9588}} ($\pm$0.0140) & 0.9475 ($\pm$0.0151) \\
                               & 10\%  & 0.0775 ($\pm$0.0044) & \underline{\textbf{0.0830}} ($\pm$0.0045) & 0.0801 ($\pm$0.0044) \\ \midrule
\multirow{2}{*}{MS Challenge}  & 100\% & 0.9143 ($\pm$0.0433) & \underline{0.9387} ($\pm$0.0083) & \underline{\textbf{0.9407}} ($\pm$0.0135) \\
                               & 10\%  & 0.0526 ($\pm$0.0091) & \underline{\textbf{0.0623}} ($\pm$0.0030) & 0.0596 ($\pm$0.0035) \\ \midrule 
\multirow{2}{*}{Android Genom} & 100\% & 0.7755 ($\pm$0.1114) & 0.8563 ($\pm$0.0941) & \underline{\textbf{0.9007}} ($\pm$0.0426) \\
                               & 10\%  & 0.0066 ($\pm$0.0052) & \underline{0.0300} ($\pm$0.0193) & \underline{\textbf{0.0326}} ($\pm$0.0182) \\ 
 \bottomrule
\end{tabular}
\end{table*}

\subsection{Datasets}
\label{sec:dataset}
We evaluate our approach using three datasets. Two malware datasets Microsoft Malware Challenge Dataset \cite{mschallengedataset2015} and Android Genome Project Dataset \cite{malgenomeproject}. The Microsoft Dataset consists of disassembled windows malware samples from 9 malware families/classes. For our evaluations, we use 10260 samples which our disassembled file parser was able to correctly process. The Android dataset consists of malicious android apps from many families/classes. In our evaluation, however, we use only classes that have at least 40 samples so as to be able to split the dataset in to training, validation and test and have enough samples. After removing the smaller classes the dataset has 986 samples. To show that our approach can be applied in other domains we evaluate our work on the MNIST Dataset\cite{mnist}, which is a dataset consisting of images of hand written digits from 0 to 9.

We extract function call graph (FCG) from the malware samples as proposed by Hassen and Chan \cite{hassen2017scalable} . In case of the android samples Android dataset we first use \cite{adagio} to extract the functions and the function instructions and then used \cite{hassen2017scalable} to extract the FCG features. For MS Challenge dataset, we reformat the FCG features as a graph adjacency matrix by taking the edge frequency features in \cite{hassen2017scalable} and rearranging  them to form an adjacency matrix. Formating the features this way allowed us to use constitutional layers on the MS Challenge dataset.

\subsection{Simulating Open Set Dataset}
\label{sec:sim_open_world}
To simulate an open world dataset for our evaluation datasets, we randomly choose $K$ number of classes from the dataset, which we will refer to as known classes in the remainder of this evaluation section, and keep only training instances from these classes in the training set. We will refer to the other classes as unknown classes. We use the open datasets created here in Sections \ref{sec:outlier_eval} and  \ref{sec:open_set_eval}.

In case of the MS Dataset and Android Dataset, first we randomly chose 6 known classes and treat set the remaining 3 as unknown classes. We then randomly select 75\% of the instances from the known classes for the training set and the remaining for the test set. We further withhold one third of the test set to serve as a validation set for hyper parameter tuning. We use only the known class instances for tuning. In these two datasets all the unknown class instances are placed into the test set. In case of the MNIST dataset, first we randomly chose 6 known classes and the remaining 4 as unknown classes. We then remove the unknown class instances from the training set. We leave the test set, which has both known and unknown class instances, as it is.  

For each of our evaluation datasets we create 3 open set datasets. We will refer to these open set datasets as OpenMNIST1, OpenMNIST2 and OpenMNIST3 for the three open set evaluation datasets created from MNIST. Similarly, we also create OpenMS1, OpenMS2, and OpenMS3 for MS Challenge dataset and OPenAndroid1, OpenAndroid2, and OpenAndroid3 for Android Genom Project dataset.

\subsection{Evaluated Approaches}
We evaluate four approaches, all implemented using Tensorflow. The first (\textit{ii}) is a network setup to be trained using ii-loss. The second (\textit{ii+ce}) is a network setup to be simultaneously trained using ii-loss and cross entropy (Section \ref{sec:iice_loss}). The third (\textit{ce}) is a network which we use to represent the baseline, is trained using cross entropy only (network setup in Figure \ref{fig:ceii_network} without the ii-loss.) The final approach is Openmax\cite{bendale2016towards} (a state-of-art algorithm), which was reimplemented based the original paper and the authors' source code to fit our evaluation framework. The authors of Openmax state that the choice of distance function Euclidean or combined Euclidean and cosine distance give similar performance in the case of their evaluation datasets \cite{bendale2016towards}. In our experiments, however, we observed that the combined Euclidean and cosine distance gives a much better performance. So we report the better result from combined Euclidean and cosine distance. 

The networks used for MS and MNIST datasets have convolution layers at the beginning followed by fully connected layers, whereas for the android dataset we use only fully connected layers. The architecture is detailed in Appendix \ref{sec:appendix_net_arch}. 

\subsection{Detecting Unknown Class Instances}
\label{sec:outlier_eval}

We start our evaluation by showing how well $outlier\_score$ (Section \ref{sec:outlier_score}) is able to identify unknown class instances. We evaluate it using 3 random open set datasets created from MS, Android and MNIST datasets as discussed in Section \ref{sec:sim_open_world}. For example, in the case of MNIST dataset, we run 10 experiments on OpenMNIST1, 10 experiments on OpenMNIST2, and 10 experiments on OpenMNIST3. We then report the average of the 30 runs. We do the same for the other two datasets.    

Table \ref{table:auc} shows the results of this evaluation. To report the results in such a way that is independent of outlier threshold, we report the area under ROC curve. This area is calculated using the outlier score and computing the true positive rate (TPR) and the false positive rate (FPR) at different thresholds. We use t-test to measure statistical significance of the difference in AUC values. Looking at the AUC up to 100\% FPR in all tree datasets, our approach \textit{ii} and \textit{ii+ce} perform significantly better(with p-value of 0.04 or less) in identifying unknown class instances than the baseline approach \textit{ce} (using only cross entropy loss.) Although AUC up to 100\% FPR gives as a full picture, in practice it is desirable to have good performance at lower false positive rates. That is is why we report AUC up to 10\% FPR. Our two approaches report a significantly better AUC than the baseline network trained to only minimize cross entropy loss.  We didn't include Openmax in this section's evaluation because it doesn't have a specific outlier score.

\subsection{Open Set Recognition}
\label{sec:open_set_eval}

\begin{figure*}[ht!]
        \centering
        \begin{subfigure}[b]{0.75\textwidth}
                \centering
                \includegraphics[scale=0.32]{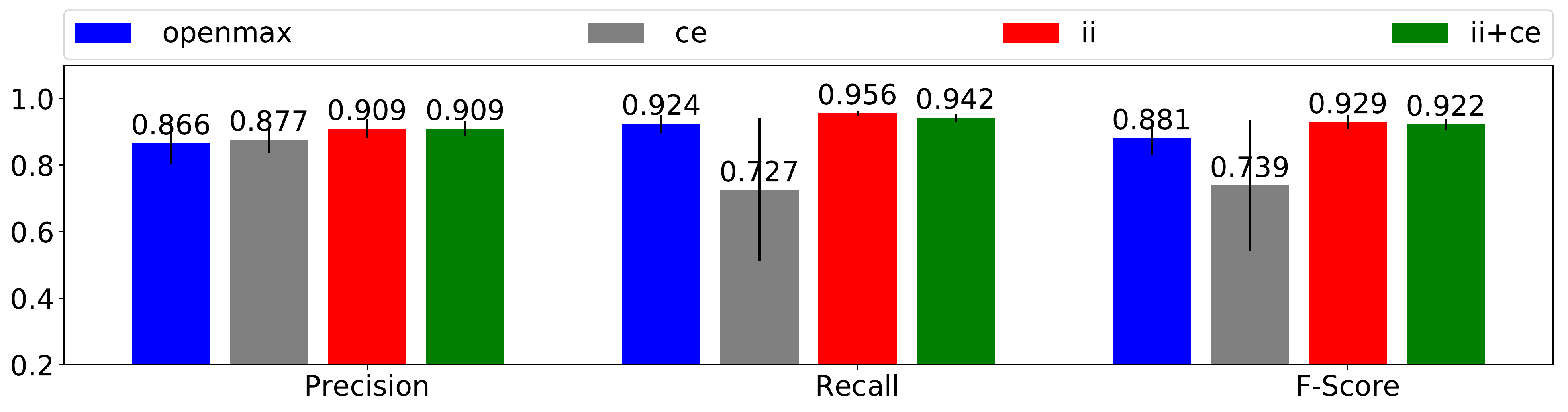}
                \caption{MNIST Dataset with 6 known and 4 unknown classes.}
                \label{fig:mnist_open_prf}
        \end{subfigure}
        
        ~ 
        \begin{subfigure}[b]{0.75\textwidth}
                \centering
                \includegraphics[scale=0.32]{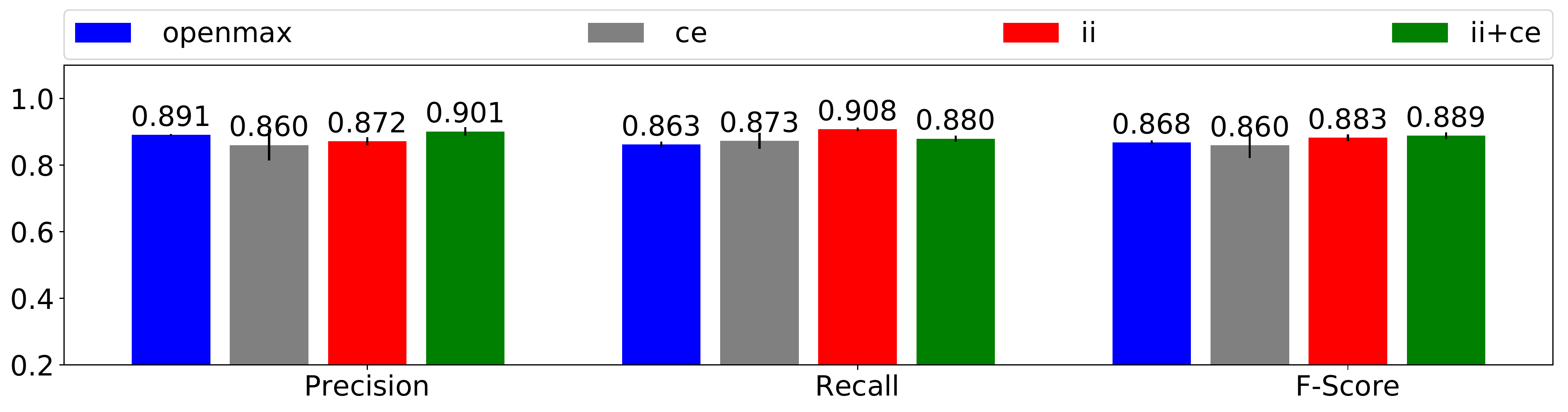}
                \caption{MS Challenge Dataset with 6 known and 3 unknown classes.}
                \label{fig:ms_open_prf}
        \end{subfigure}
        
        \begin{subfigure}[b]{0.75\textwidth}
                \centering
                \includegraphics[scale=0.32]{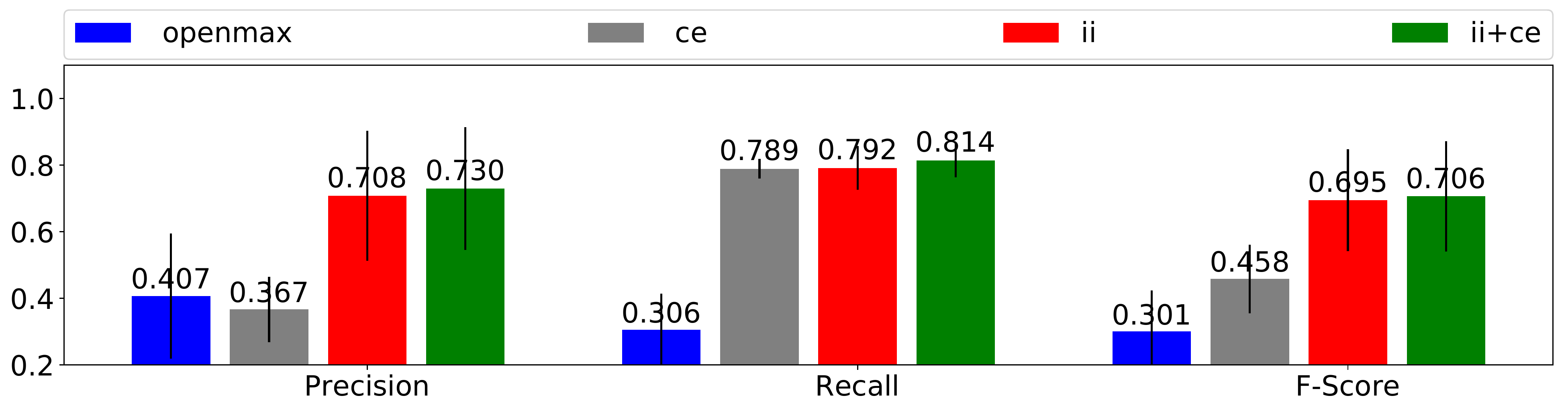}
                \caption{Android Genom Dataset with 6 known and 3 unknown classes.}
                \label{fig:android_open_prf}
        \end{subfigure}
        \caption{Average Precision, Recall and F-Score of 30 Runs. The metric calculations included the average of $K$ known class labels and the one $unknown$ class label further averaged over 30 experiment runs. The results are from Openmax, a baseline network trained on cross entropy loss only (\textit{ce}), a network trained on ii-loss only (\textit{ii}), and a network trained on combination of ii-loss and cross entropy (\textit{ii+ce}).}
        \label{fig:open_prf}
\end{figure*}

When the proposed approach is used for open set recognition, the final prediction is a class label, which can be with one of the $K$ known class labels if the a test instances has an outlier score less than a threshold value (Section \ref{sec:threshold}) or it can be an ``$unknown$" label if the instance has an outlier score greater than the threshold. In addition to the three approaches evaluated in the previous section, we also include Openmax \cite{bendale2016towards} in these evaluations because it gives final class label predictions.

We use average precision, recall and f-score metrics to evaluate open set recognition performance and t-test for statistical significance. Precision, recall and f-score are first calculated for each of the $K$ known classes labels and the one ``$unknown$" label. Then the average overall the $K + 1$ classes is calculated. Using the same experimental setup as Section \ref{sec:outlier_eval} (i.e. using 3 random open set datasets created from MS, Android and MNIST datasets as discussed in Section \ref{sec:sim_open_world}.), we report the result of the average  precision, recall and f-score, averaged across all class labels and 30 experiment runs in Figure \ref{fig:open_prf}.

On all three datasets the \textit{ii} and \textit{ii+ce} networks gives significantly better f-score compared to the other two configurations (with p-value of 0.0002 or less). In case of the Android dataset, all networks perform lower compared to the other two datasets. We attribute this to the small number of samples in the Android datasets. The dataset is also imbalanced with a number of classes only having less than 60 samples.

\subsection{Closed Set Classification}
In this section we would like to show that on a closed dataset, a network trained using ii-loss performs comparably to the same network trained using cross entropy loss. For closed set classification, all the classes in the dataset are used for both training and test. For MS and Android datasets we randomly divide the datasets into training, validation, and test and report the results on the test set. The MNIST dataset is already divided into training, validation and test. 

On closed MNIST dataset, a network trained with cross entropy achieved, a 10-run average classification accuracy of 99.42\%. The same network trained using ii-loss achieved an average accuracy of 99.31\%. The network trained only on cross entropy gives better performance than the network trained on ii-loss. We acknowledge that both results are not state-of-art as we are using simple network architectures. The main goal of these experiments is to show that the ii-loss trained network can give comparable results to a cross entropy trained network. On the Android dataset the network trained on a cross entropy gets an average classification accuracy of 93.10\% while ii-loss records 92.68\%, but the difference is not significant (with p-value at 0.43).

In Section \ref{sec:iice_loss}, we proposed training a network on both ii-loss cross entropy loss in an effort to get lower classification error. The results from our experiments using such a network for closed MNIST dataset give an average classification accuracy of 99.40\%. This result makes it comparable to the performance of the same network trained using cross entropy only (with p-value of 0.22).   

\subsection{Discussions}
\label{discussions}

As discussed in Section \ref{sec:related_works} the limitations of openmax are 1) it does not use a loss function that directly incentivizes projecting class instances around the mean class activation vector and 2) the distance function used by openmax is not necessarily the right distance function for final activation vector space. We addressed these limitations by training a neural network with a loss function that explicitly encourages the two properties in Section \ref{sec:overview}. In addition, we use the same distance function during training and test. As a result, we observe in Section \ref{sec:open_set_eval} that our two proposed approaches perform better in open set recognition.

\begin{figure*}[ht]
        \centering
        \begin{subfigure}[b]{0.53\textwidth}
                \centering
                \hspace*{-1.5cm}\includegraphics[scale=0.4]{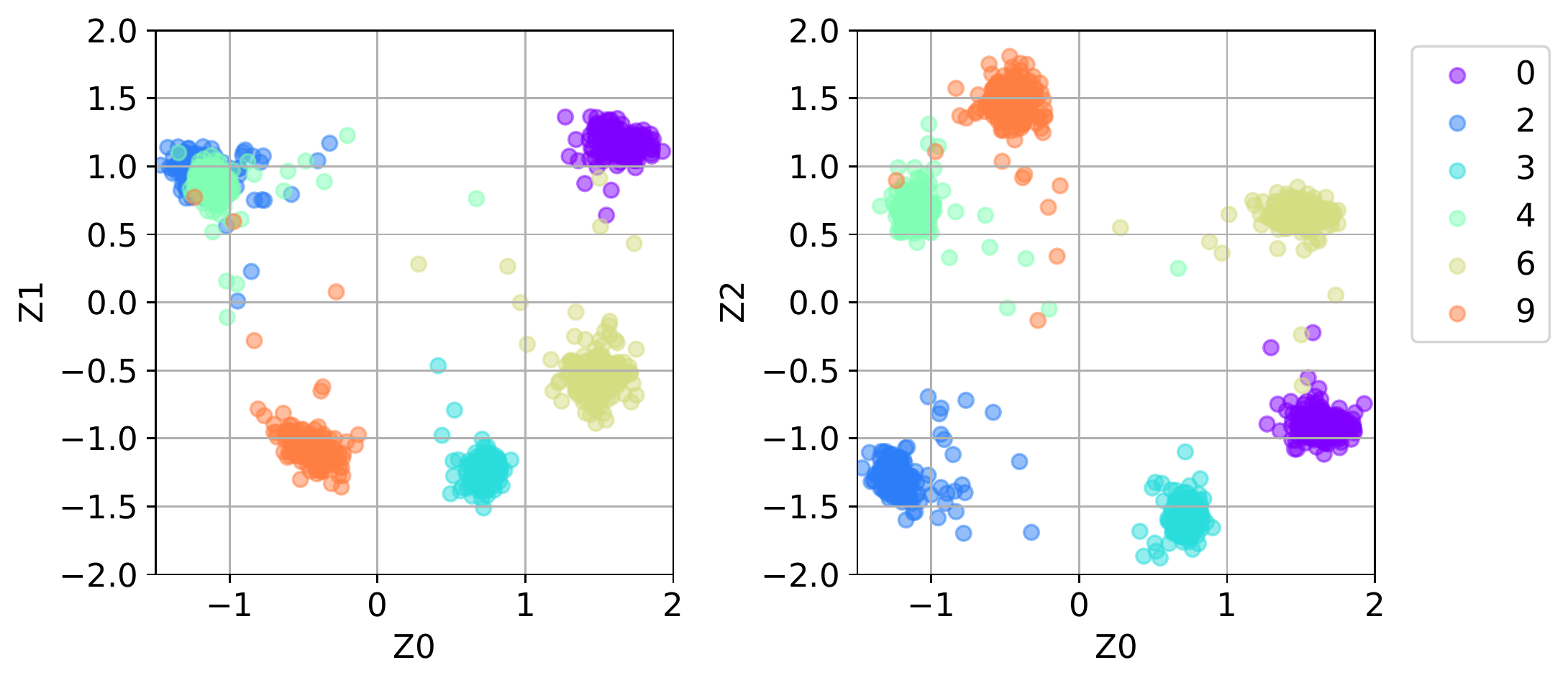}
                \caption{ii-loss}
                \label{fig:z_plot_known}
        \end{subfigure}
        \begin{subfigure}[b]{0.46\textwidth}
                \centering
                \hspace*{-.61cm}\includegraphics[scale=0.4]{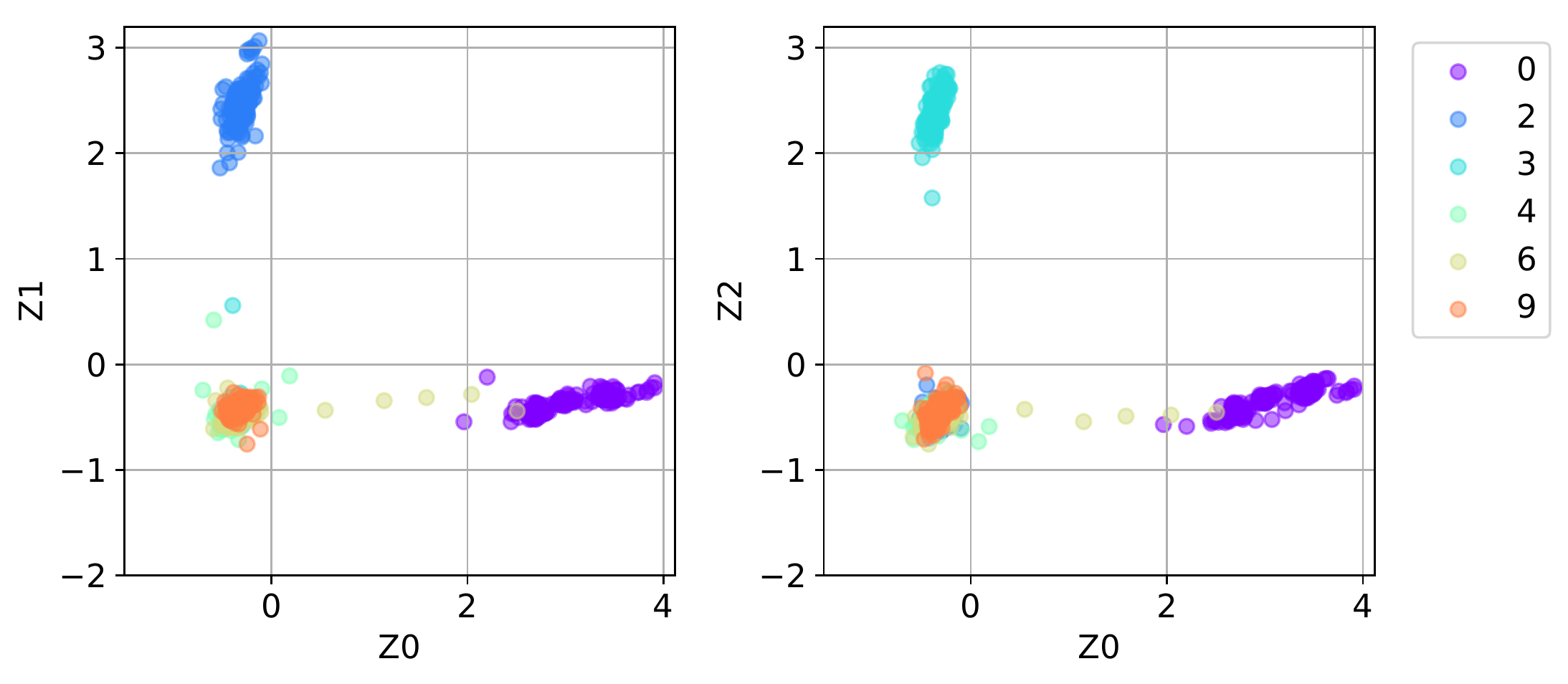}
                \caption{openmax}
                \label{fig:z_plot_known_openmax}
        \end{subfigure}
        
        \begin{subfigure}[b]{0.53\textwidth}
                \centering
                \hspace*{-1cm}\includegraphics[scale=0.4]{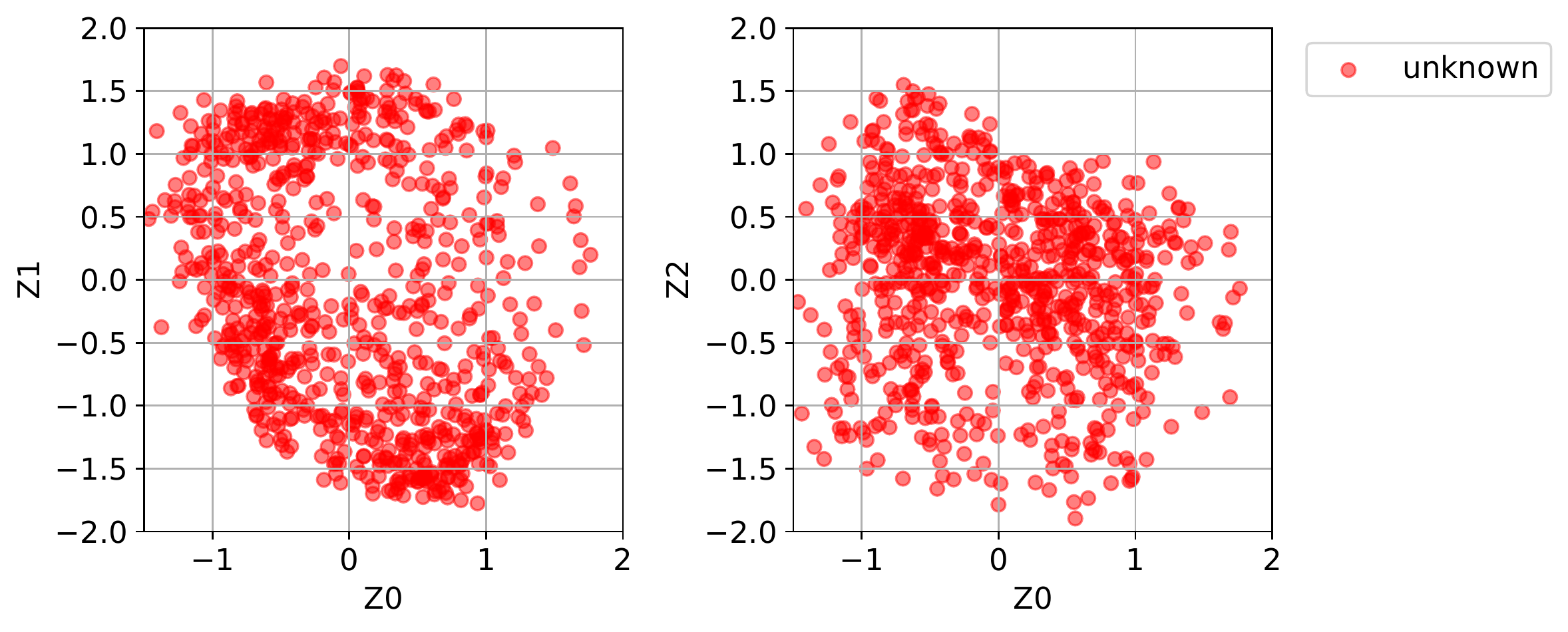}
                \caption{ii-loss}
                \label{fig:z_plot_unknown}
        \end{subfigure}
        \begin{subfigure}[b]{0.45\textwidth}
                \centering
                \hspace*{-.7cm}\includegraphics[scale=0.4]{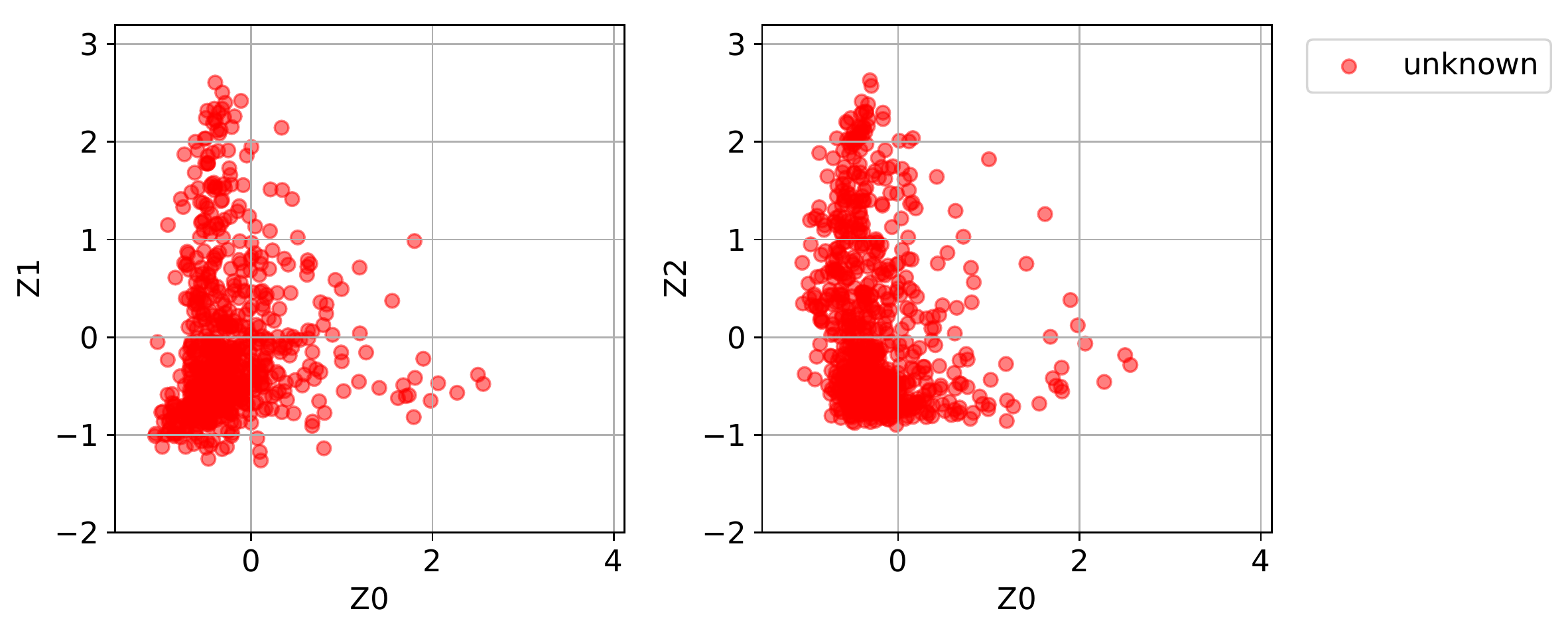}
                \caption{openmax}
                \label{fig:z_plot_unknown_openmax}
        \end{subfigure}
        \caption{The z-layer projection of (a, b) known and (c, d) unknown class instances from test set of MNIST dataset. The labels 0,2,3,4,6,9 represent the known classes while the label ``unknown" represents the unknown classes.}
        \label{fig:z_plot}
\end{figure*}

Figure \ref{fig:z_plot} provides evidence on how our network projects unknown class instances in the space between the known classes. In the figure the z-layer projection of 2000 random test instances of an open set dataset created from MNIST with 6 known and 4 unknown classes. The class labels 0, 2, 3, 4, 6, and 9 in the figure represent the 6 known classes while the ``unknown" label represents all the unknown classes. The network with ii-loss is setup to have a z-layer dimension of 6, and the figure shows a 2D plot of dimension (z0,z1), (z0,z2). The Openmax network also has a similar network architecture and last layer dimension of 6. In case of ii-loss based projection, the instances from the known classes (Figure \ref{fig:z_plot_known}) are projected close to their respective class while the unknown class instances (Figure \ref{fig:z_plot_unknown}) are projected, for the most part, in the region between the classes. It is this property that allows us to use the distance from the class mean as an outlier score to perform open set recognition. In case of openmax, Figures \ref{fig:z_plot_known_openmax} and \ref{fig:z_plot_unknown_openmax}, the unknown class instances do not fully occupy the open space between the known classes. In openmax, most instances are projected along the axis, this is because of the one-hot encoding induced by cross entropy loss. So compared to openmax, ii-loss appears to better utilize space ``among" the classes. 

\begin{figure}[ht]
        \centering
        \includegraphics[scale=0.34]{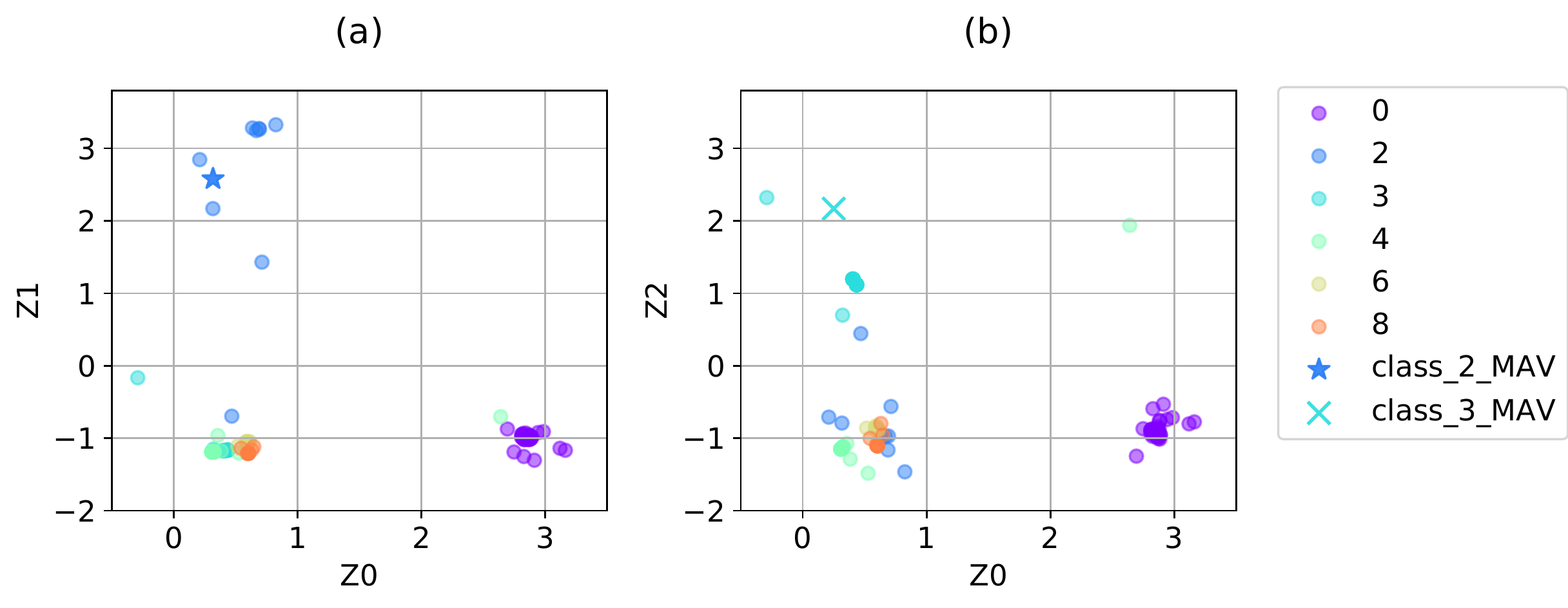}
        \caption{Projections of  Android dataset known class test instances from final activation layer of Openmax.}
        \label{fig:z_plot_android_known}
\end{figure}

\begin{figure}[ht]
        \centering
        \begin{subfigure}[b]{0.22\textwidth}
                \centering
                \hspace*{-0.8cm}\includegraphics[scale=0.45]{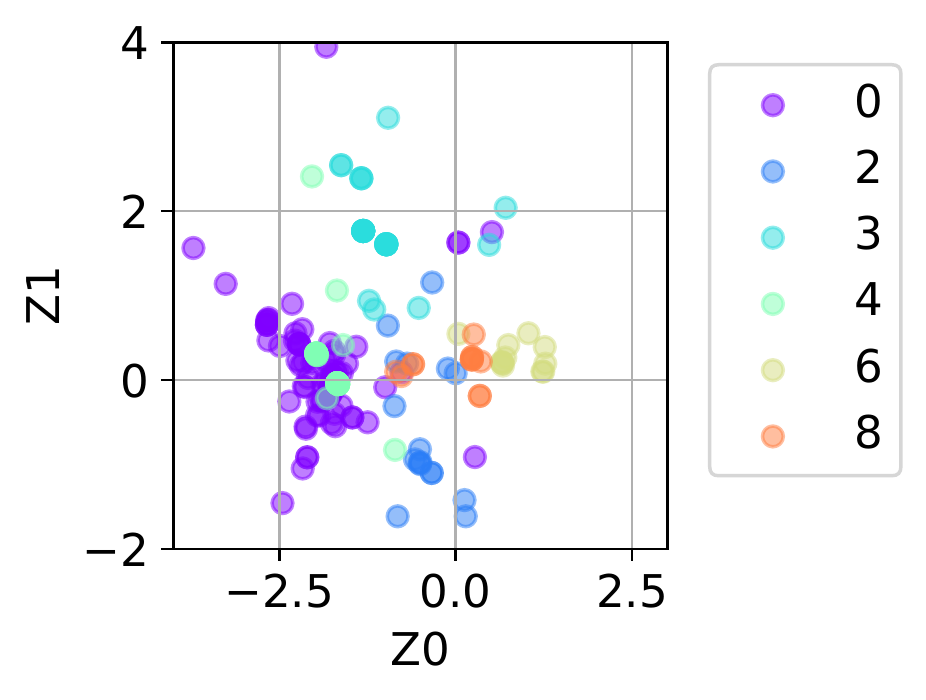}
                \caption{}
                \label{fig:z_plot_android_ce_known}
        \end{subfigure}
        \begin{subfigure}[b]{0.23\textwidth}
                \centering
                \hspace*{-0.2cm}\includegraphics[scale=0.45]{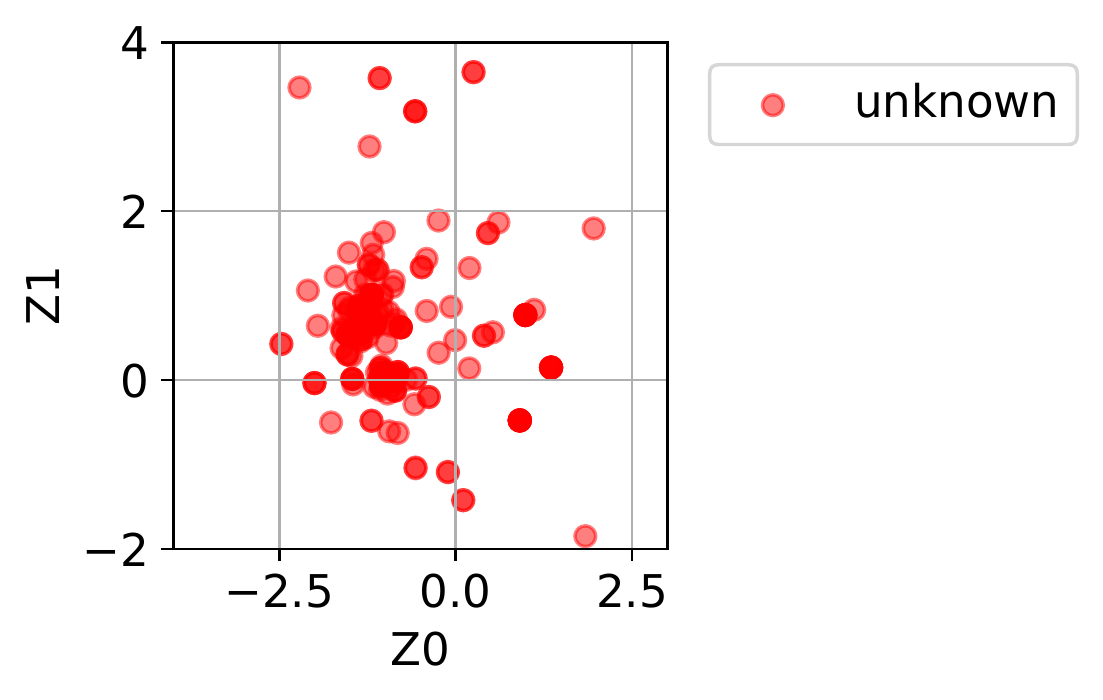}
                \caption{}
                \label{fig:z_plot_android_ce_unknown}
        \end{subfigure}
        \caption{Projections of  Android dataset (a) known class and (b) unknown class test instances from z-layer of a network trained with only cross entropy.}
        \label{fig:z_plot_android_ce}
\end{figure}

Performance of Openmax is especially low in case of the Android dataset because of low recall on known classes with small number training instances. The low recall was caused by test instances from the smaller classes being projected further away from the class's mean activation vector (MAV). For example, in Figure \ref{fig:z_plot_android_known}a we see that test instances of class 2 are further away from the MAV of class 2 (marked by '$\star$'). As a result, these test instances are predicted as unknown. Similarly, in Figure \ref{fig:z_plot_android_known}b instances of class 3 are far away from the MAV of class 3(marked by 'X'). Performance of ce is also low for Android dataset. This was because unknown class instances were projected close to the known classes (Figure \ref{fig:z_plot_android_ce}) resulting them being labeled as known classes. In turn resulting in lower precision score for the known classes.

\begin{figure}[ht]
        \centering
        \begin{subfigure}[b]{0.22\textwidth}
                \centering
                \includegraphics[scale=0.53]{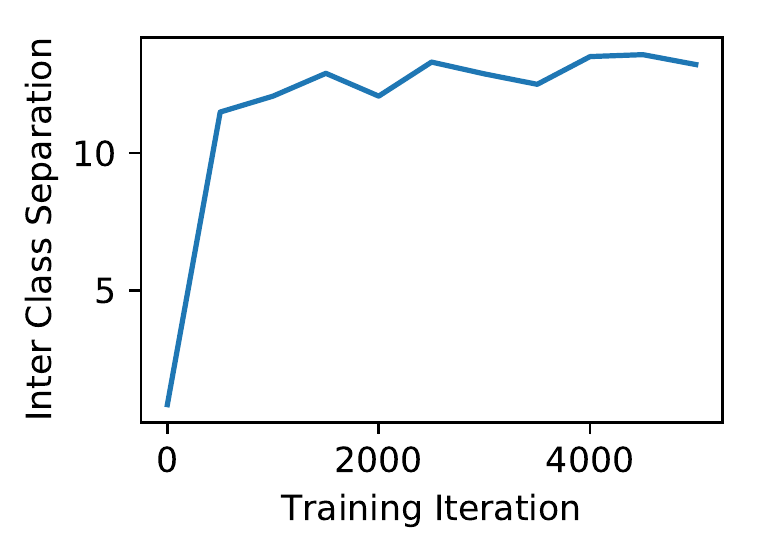}
                \caption{}
                \label{fig:yes_batch_norm}
        \end{subfigure}
        \begin{subfigure}[b]{0.23\textwidth}
                \centering
                \includegraphics[scale=0.53]{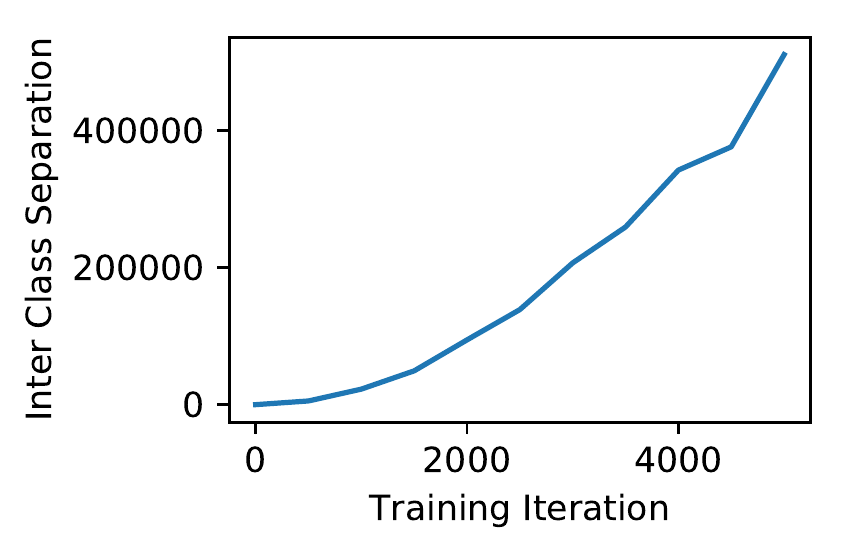}
                \caption{}
                \label{fig:no_batch_norm}
        \end{subfigure}
        \caption{Inter class separation for networks trained (a) with batch normalization used in all layers and (b) without batch normalization at the z-layer.}
        \label{fig:batch_norm}
\end{figure}

On the topic of the z-layer, one question the reader might have is how to decide on the dimension of this layer. Our suggestion is to chose this experimentally as is the case for other neural network hyper-parameters. In our case, we tuned the z-layer dimension together with other hyper-parameters for experiments in Sections \ref{sec:outlier_eval} and \ref{sec:open_set_eval} based on the closed set performance on the known class instances in validation set. Based on the results we decided to keep z-layer dimension equal to the number of classes as we didn't see significant improvement from increasing it.

In our experiments, we have observed batch normalization \cite{ioffe2015batch} to be extremely important when using ii-loss. Because batch normalization fixes the mean and variance of a layer, it bounds the output of our z-layer in a certain hypercube, in turn preventing the $inter\_separation$ term in ii-loss from increasing indefinitely. This is evident in Figures \ref{fig:yes_batch_norm} and \ref{fig:no_batch_norm}. Figure \ref{fig:yes_batch_norm} shows the  $inter\_separation$ of the network where batch normalization used in all layers including the z-layer where $inter\_separation$ increases in the beginning but levels off. Whereas when batch normalization is not used in the z-layer the $inter\_separation$ term keeps on increasing as seen in Figure \ref{fig:no_batch_norm}.

\begin{figure}[ht]
        \centering
        \begin{subfigure}[b]{0.53\textwidth}
                \centering
                \hspace*{-1.5cm}\includegraphics[scale=0.34]{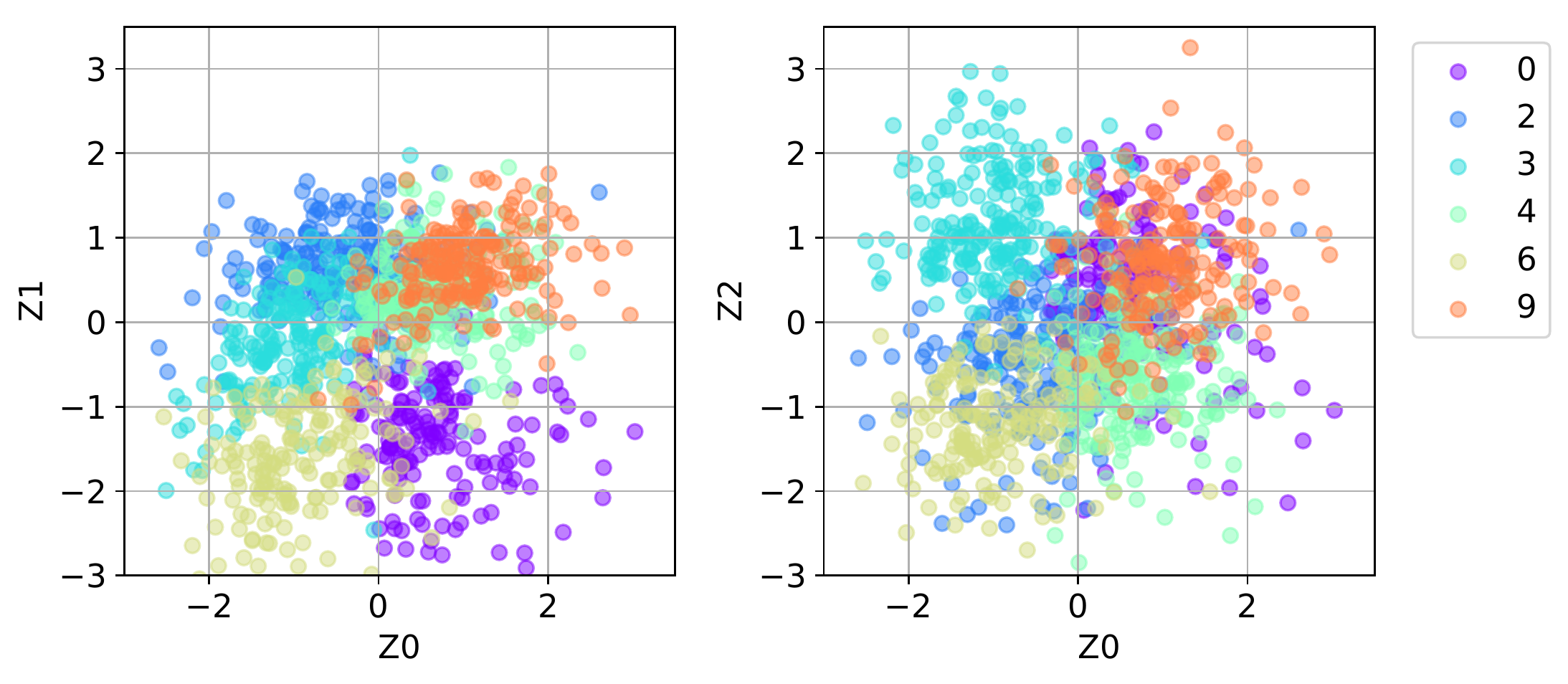}
                \caption{}
                \label{fig:z_plot_known_autoencoder}
        \end{subfigure}
        
        \begin{subfigure}[b]{0.53\textwidth}
                \centering
                \hspace*{-1cm}\includegraphics[scale=0.34]{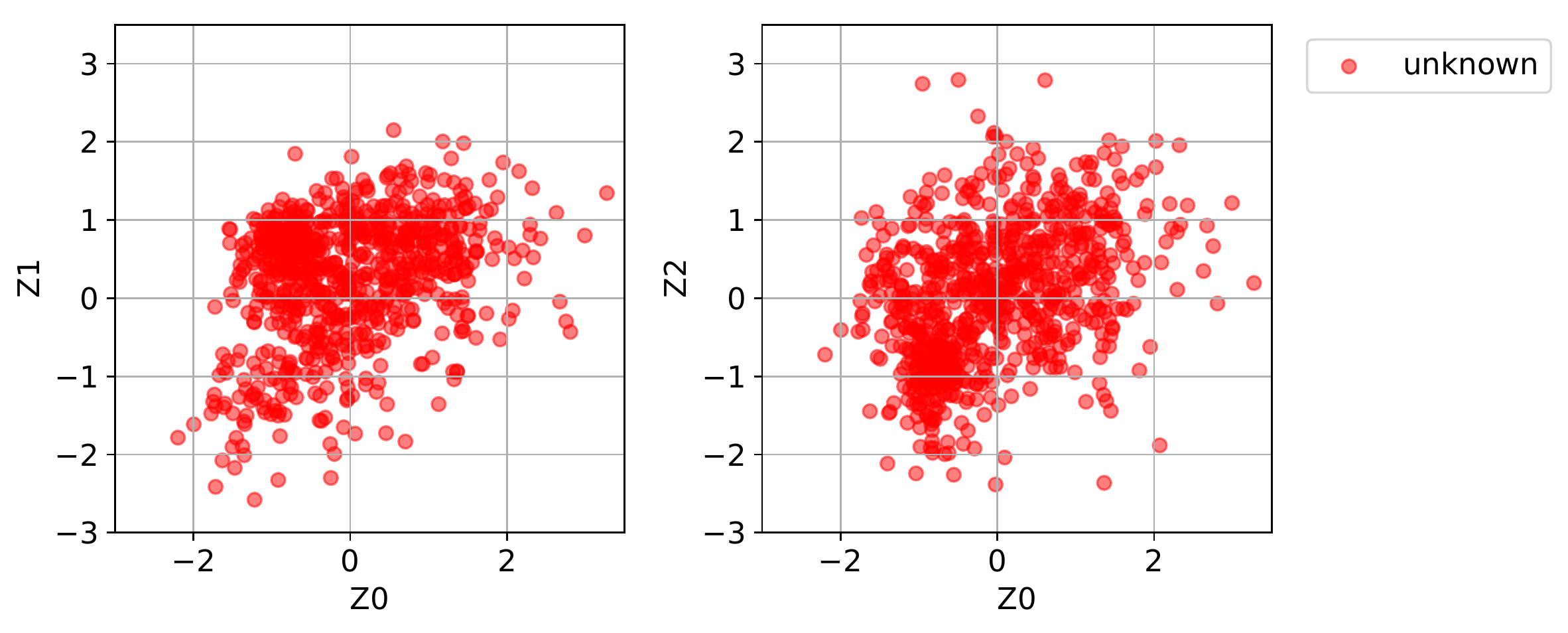}
                \caption{}
                \label{fig:z_plot_unknown_autoencoder}
        \end{subfigure}
        \caption{Projections of (a) known and (b) unknown class instances using the hidden layer of an Autoencoder. The labels 0,2,3,4,6,9 represent the known classes while the label ``unknown" represents the unknown classes.}
        \label{fig:z_plot_autoencoder}
\end{figure}

Autoencoders can also be considered as another way to learn a representation. However, autoencoders do not try to achieve properties P1 and P2 in Section \ref{sec:overview}. One of the reasons is auto encoder training is unsupervised. Another reason is because non-regularized autoencoders fracture the manifold into different domains resulting in the representation of instances from the same class being further apart \cite{makhzani2015adversarial}. Therefore, the representation learned does not help in discriminating known class instances from unknown class instances. Figure \ref{fig:z_plot_autoencoder} shows the output of an encoder in an autoencoder trained on known class instances and then used to project both known and unknown class instances. For the projection in autoencoder, the known classes are not well separated and outliers get projected to roughly the same area as the known classes.

In Section \ref{sec:dataset} we mentioned that we used function call graph (FCG) feature for the malware dataset. We also mentioned that in case of the MS Challenge dataset we reformatted the FCG features proposed in \cite{hassen2017scalable} to form a $(63, 63)$ adjacency matrix  representation of the graph. We feed this matrix as an input to the convolutional network with a (4,4) kernel. Such kernel shape makes sense when it comes to image input because in images proximity of pixels hold important information. However, it is not obvious to us how nearby cells in a graph adjacency matrix hold meaning full information. We tried different kernel shapes, for example taking an entire row of the matrix at once (because a row of the matrix represent a single nodes outgoing edge weights). However the simple (4,4) giving better close set performance.  

\section{Conclusion}
We presented an approach for learning a neural network based representation that projects instances of the same class closer together while projecting instances of the different classes further apart. Our empirical evaluation shows that the two properties lead to larger spaces among classes for instances of unknown classes to occupy, hence facilitating open set recognition. We compared our proposed approach with a baseline network trained to minimize a cross entropy loss and with Openmax (a state-of-art neural network based open set recognition approach). We evaluated the approaches on datasets of malware samples and images and observed that our proposed approach achieves statistically significant improvement.  

We proposed a simple threshold estimation technique, Section \ref{sec:threshold}. However, there is room to explore a more robust way to estimate the threshold. We leave this for future work.

\appendix
\section{Evaluation Network Architectures}
\label{sec:appendix_net_arch}

We evaluate 4 networks: ii, ce, ii+ce, and Openmax. All four networks have the same architecture upto the fully connected z-layer. In case of the MNIST dataset, the input images are of size (28,28) and are padded to get an input layer size (32,32) with 1 channel. Following the input layer are 2 non-linear convolutional layers with 32 and 64 unites (filters) which have a kernel size of (4,4) with a (1,1) strides and SAME padding. The network also has max polling layers with kernel size of (3,3), strides of (2,2), and SAME padding after each convolutional layer. Two fully connected non-linear layers with 256 and 128 units follow the second max pooling layer. Then a linear z-layer with dimension of 6 follows the fully connected layers. In the case of ii+ce and ce networks, the output the z-layer are fed to an additional linear layer of dimension 6 which is then given to a softmax function. We use Relu activation function for all the non-linear layers. Batch normalization is used though out all the layers. We also use Dropout with keep probability of 0.2 for the fully connected layers. Adam optimizer with learning rate of 0.001, beta1 of 0.5, and beta2 of 0.999 is used to train our networks for 5000 iterations. In case of the Openmax network, the output of the z-layer are directly fed to a softmax layer. Similar to the Openmax paper we use a distance that is a weighted combination of normalized Euclidean and cosine distances. For the ce, ii, and ii+ce we use contamination ratio of 0.01 for the threshold selection.     

The open set experiments for MS Challenge dataset also used a similar architectures as the four networks used for MNIST dataset with the following differences. The input layer size MS Challenge dataset is (67,67) with 1 channel after padding the original input of (63,63). Instead of the two fully connected non-linear layers, we use one fully connected layer with 256 units. We use dropout in the fully connected layer with keep probability of 0.9. Finally the network was trained using Adam optimizer  with 0.001 learning rate, 0.9 beta1 and 0.999 beta2.

We do not use a convolutional network for the Android dataset open set experiments. We use a network with one fully connected layer of 64 units. This is followed by a z-layer with dimension of 6. For ii+ce and ce networks we further add a linear layer with dimension of 6 and the output of this layer is fed to a softmax layer. In case of Openmax the output of the z-layer is directly fed to the softmax layer. For Openmax we use a distance that is a weighted combination of normalized Euclidean and cosine distances. We use Relu activation function for all the non linear layers. We used batch normalization for all layers. We also used Dropout with keep probability of 0.9 for the fully connected layers. We used Adam optimizer with learning rate of 0.1 and first momentem of 0.9 to train our networks for 10000 iterations. For the ce, ii, and ii+ce we use contamination ratio of 0.01 for the threshold selection.

The closed set experiments use the same set up as the open set experiments with the only difference coming from the dimension of the z-layer. For the MNIST dataset we used z dimension of 10. For the MS and Android datasets we use z dimension of 9.

\begin{acks}

The experiments in this work were run on resources supported by the National Science Foundation under Grant No. CNS 09-23050. We would also like to thank Abhijit Bendale for sharing his source code to use as a reference for our implementation of Openmax.

\end{acks}

\bibliographystyle{abbrv}
\bibliography{open_nn.bib}

\begin{thebibliography}{10}

\bibitem{adagio}
Adagio.
\newblock https://github.com/hgascon/adagio.

\bibitem{malgenomeproject}
Android malware genome project.
\newblock http://www.malgenomeproject.org/.

\bibitem{mnist}
Mnist hand written digit dataset.
\newblock http://yann.lecun.com/exdb/mnist/.

\bibitem{mschallengedataset2015}
Microsoft malware classification challenge (big 2015).
\newblock https://www.kaggle.com/c/malware-classification, 2015.
\newblock [Online; accessed 27-April-2015].

\bibitem{bendale2015towards}
A.~Bendale and T.~Boult.
\newblock Towards open world recognition.
\newblock In {\em Proceedings of the IEEE Conference on Computer Vision and
  Pattern Recognition}, pages 1893--1902, 2015.

\bibitem{bendale2016towards}
A.~Bendale and T.~E. Boult.
\newblock Towards open set deep networks.
\newblock In {\em Proceedings of the IEEE Conference on Computer Vision and
  Pattern Recognition}, pages 1563--1572, 2016.

\bibitem{bodesheim2015local}
P.~Bodesheim, A.~Freytag, E.~Rodner, and J.~Denzler.
\newblock Local novelty detection in multi-class recognition problems.
\newblock In {\em Applications of Computer Vision (WACV), 2015 IEEE Winter
  Conference on}, pages 813--820. IEEE, 2015.

\bibitem{bodesheim2013kernel}
P.~Bodesheim, A.~Freytag, E.~Rodner, M.~Kemmler, and J.~Denzler.
\newblock Kernel null space methods for novelty detection.
\newblock In {\em Proceedings of the IEEE Conference on Computer Vision and
  Pattern Recognition}, pages 3374--3381, 2013.

\bibitem{cardoso2015bounded}
D.~O. Cardoso, F.~Fran{\c{c}}a, and J.~Gama.
\newblock A bounded neural network for open set recognition.
\newblock In {\em Neural Networks (IJCNN), 2015 International Joint Conference
  on}, pages 1--7. IEEE, 2015.

\bibitem{da2014learning}
Q.~Da, Y.~Yu, and Z.-H. Zhou.
\newblock Learning with augmented class by exploiting unlabeled data.
\newblock In {\em Twenty-Eighth AAAI Conference on Artificial Intelligence},
  2014.

\bibitem{dietterich2017steps}
T.~G. Dietterich.
\newblock Steps toward robust artificial intelligence.
\newblock {\em AI Magazine}, 38(3):3--24, 2017.

\bibitem{ge2017generative}
Z.~Ge, S.~Demyanov, Z.~Chen, and R.~Garnavi.
\newblock Generative openmax for multi-class open set classification.
\newblock {\em arXiv preprint arXiv:1707.07418}, 2017.

\bibitem{goodfellow2014generative}
I.~Goodfellow, J.~Pouget-Abadie, M.~Mirza, B.~Xu, D.~Warde-Farley, S.~Ozair,
  A.~Courville, and Y.~Bengio.
\newblock Generative adversarial nets.
\newblock In {\em Advances in neural information processing systems}, pages
  2672--2680, 2014.

\bibitem{hassen2017scalable}
M.~Hassen and P.~K. Chan.
\newblock Scalable function call graph-based malware classification.
\newblock In {\em 7th Conference on Data and Application Security and Privacy},
  pages 239--248. ACM, 2017.

\bibitem{ioffe2015batch}
S.~Ioffe and C.~Szegedy.
\newblock Batch normalization: Accelerating deep network training by reducing
  internal covariate shift.
\newblock In {\em International Conference on Machine Learning}, pages
  448--456, 2015.

\bibitem{Jain2014}
L.~P. Jain, W.~J. Scheirer, and T.~E. Boult.
\newblock {Multi-Class Open Set Recognition Using Probability of Inclusion}.
\newblock pages 393--409, 2014.

\bibitem{makhzani2015adversarial}
A.~Makhzani, J.~Shlens, N.~Jaitly, I.~Goodfellow, and B.~Frey.
\newblock Adversarial autoencoders.
\newblock {\em arXiv preprint arXiv:1511.05644}, 2015.

\bibitem{ortiz2014face}
E.~G. Ortiz and B.~C. Becker.
\newblock Face recognition for web-scale datasets.
\newblock {\em Computer Vision and Image Understanding}, 118:153--170, 2014.

\bibitem{radford2015unsupervised}
A.~Radford, L.~Metz, and S.~Chintala.
\newblock Unsupervised representation learning with deep convolutional
  generative adversarial networks.
\newblock {\em arXiv preprint arXiv:1511.06434}, 2015.

\bibitem{rieck2011automatic}
K.~Rieck, P.~Trinius, C.~Willems, and T.~Holz.
\newblock Automatic analysis of malware behavior using machine learning.
\newblock {\em Journal of Computer Security}, 19(4):639--668, 2011.

\bibitem{rudd2017survey}
E.~Rudd, A.~Rozsa, M.~Gunther, and T.~Boult.
\newblock A survey of stealth malware: Attacks, mitigation measures, and steps
  toward autonomous open world solutions.
\newblock {\em IEEE Communications Surveys \& Tutorials}, 2017.

\bibitem{scheirer2013toward}
W.~J. Scheirer, A.~de~Rezende~Rocha, A.~Sapkota, and T.~E. Boult.
\newblock Toward open set recognition.
\newblock {\em IEEE Transactions on Pattern Analysis and Machine Intelligence},
  35(7):1757--1772, 2013.

\end{thebibliography}

\end{document}